\documentclass[conference]{IEEEtran}
\IEEEoverridecommandlockouts
\newcommand{\CLASSINPUTtoptextmargin}{2.4cm}
\usepackage{cite}
\usepackage{amsmath,amssymb,amsfonts}
\usepackage{algorithmic}
\usepackage{graphicx}
\usepackage{textcomp}
\usepackage{xcolor}
\usepackage{amsmath,graphicx}
\usepackage{color, colortbl}
\usepackage{float}
\usepackage{tabularx}
\usepackage{booktabs}
\usepackage{url}
\usepackage{mathtools}
\usepackage{dblfloatfix}

\definecolor{Gray}{gray}{0.9}
\def\BibTeX{{\rm B\kern-.05em{\sc i\kern-.025em b}\kern-.08em
		T\kern-.1667em\lower.7ex\hbox{E}\kern-.125emX}}
\begin{document}
	
	\title{VoiceMoji: A Novel On-Device Pipeline for Seamless Emoji Insertion in Dictation
		\thanks{978-1-6654-4175-9/21/\$31.00 ©2021 IEEE}}

	\author{
		\IEEEauthorblockN{Sumit Kumar}
		\IEEEauthorblockA{\textit{Samsung R\&D Institute} \\
			Bangalore, India \\
			sumit.kr@samsung.com}
		\and
		\IEEEauthorblockN{Harichandana B S S}
		\IEEEauthorblockA{\textit{Samsung R\&D Institute} \\
			Bangalore, India \\
			hari.ss@samsung.com}
		\and
		\IEEEauthorblockN{Himanshu Arora}
		\IEEEauthorblockA{\textit{Samsung R\&D Institute} \\
			Bangalore, India \\
			him.arora@samsung.com}
		
	}
	
	\maketitle

	\begin{abstract}
		Most of the speech recognition systems recover only words in the speech and fail to capture emotions. Users have to manually add emoji(s) in text for adding tone and making communication fun. Though there is much work done on punctuation addition on transcribed speech, the area of emotion addition is untouched. In this paper, we propose a novel on-device pipeline to enrich the voice input experience. It involves, given a blob of transcribed text, intelligently processing and identifying structure where emoji insertion makes sense. Moreover, it includes semantic text analysis to predict emoji for each of the sub-parts for which we propose a novel architecture Attention-based Char Aware (ACA) LSTM which handles Out-Of-Vocabulary (OOV) words as well. All these tasks are executed completely on-device and hence can aid on-device dictation systems. To the best of our knowledge, this is the first work that shows how to add emoji(s) in the transcribed text. We demonstrate that our components achieve comparable results to previous neural approaches for punctuation addition and emoji prediction with 80\% fewer parameters. Overall, our proposed model has a very small memory footprint of a mere 4MB to suit on-device deployment.
	\end{abstract}
	

	\section{Introduction}
	\label{sec:intro}
	
	Speech recognition systems have been around for more than five decades with the latest systems achieving Word Error Rates (WER) of 5.5\% \cite{saon2017english} \cite{xiong2016achieving}, owing to the advent of deep learning. Due to existing data security and privacy concerns in cloud-based ASR systems, a clear shift in preference towards on-device deployment of the state-of-the-art Automated Speech Recognition (ASR) models is emerging \cite{mehrotra2020iterative}.
	
	\begin{figure}[ht]
	\centering
	\includegraphics[width=\linewidth, scale=0.8]{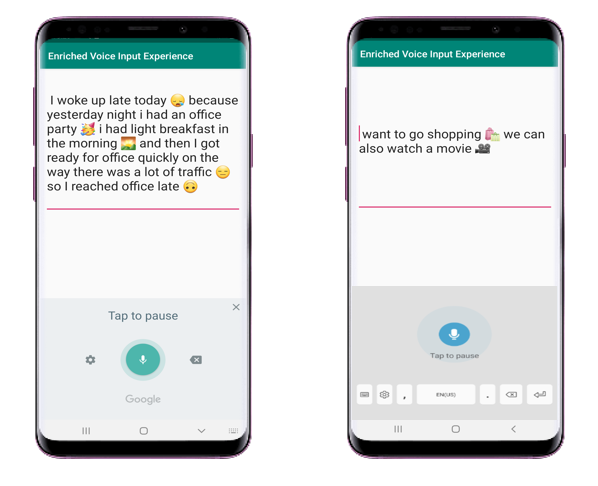}
	\caption{Voicemoji output of Sample voice input from Google and Bixby voice assistants.}
	\label{fig:screen}
\end{figure}
\begin{figure*}[!b]
	\centering
	\includegraphics[width=\linewidth]{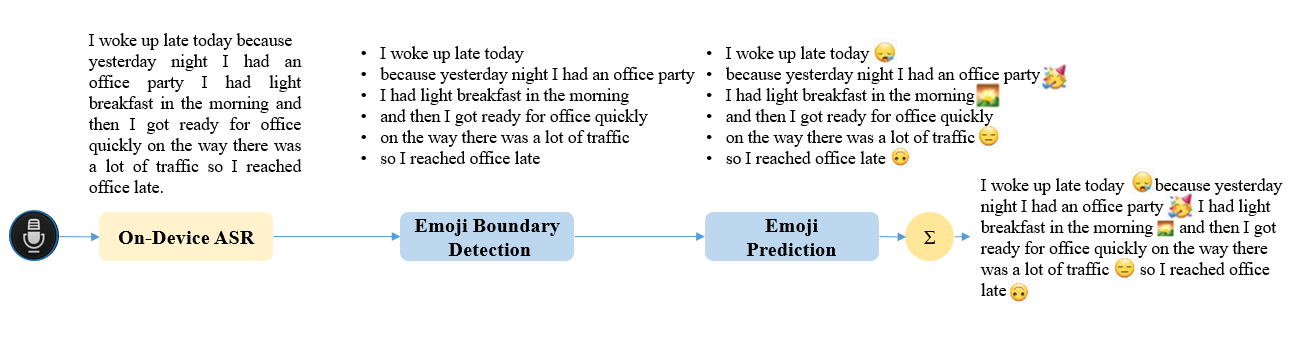}
	\caption{Processing Pipeline for Emoji insertion in dictation.}
	\label{fig:process pipeline}
\end{figure*}
	Gartner predicts that by 2022, 80\% of smartphones shipped will have on-device AI capabilities \cite{gartner}. As per a recent survey by Perficient \cite{perficient} on mobile voice usage, over 19 percent of users prefer to dictate messages. However, while the world has become increasingly multimodal with chats now involving various modalities such as emojis, stickers, etc., speech recognition systems only transcribe words in the speech and remain apathetic towards such modalities. To the best of our knowledge, existing methods of adding emoji(s) to transcribed text rely on either the user interface (UI) based methods or explicit mention of emoji e.g. ``fire emoji"  and not on automatic insertion of emoji(s). The latter method of dictating emoji is unnatural and inadequate. Also, the user may not remember the keywords to include the specific type of emojis as per the context. There has been work done on adding multiple emojis to text directly \cite{jiang2020automatic} but not using transcribed text from audio input which is our main focus. In this paper, we show a system for transcribing text from the speech that automatically adds emoji(s) to the transcribed text, some samples of which are shown in Figure \ref{fig:screen}.
	
	Our system consists of a text processing module and a text understanding module. In the text processing module, we detect text boundaries and treat them as a punctuation prediction and sentence sub-part detection task. In the last two decades, a substantial amount of research has been done on punctuation prediction and related tasks like sentence boundary detection, and sentence unit detection. But most work use only lexical features as written data with punctuation is abundant compared to audio data with processed prosodic features.  Also, prosody cues are inconsistent and may vary from person to person \cite{adami2003modeling}. Due to the above reasons, we too refrain from using prosody cues in our model.
	Moreover, prosody processing differs for different Speech To Text (STT) engines. The solution we present can work for all STT engines as it works on the textual output of the engine.

	In the text understanding module, we predict emoji(s) for the sentence sub-part(s) and render an emoji-rich transcribed text. Emojis are used everywhere from social media to chatting apps. While speech recognition systems reduce the typing need of a user, they do not provide the user with a completely hands-off experience that these systems were originally intended for. Our system serves the user's need of adding emotion to text, thus creating a truly hands-off experience for the user. Our contributions can be summarized as follows: 
	
	\begin{itemize}
		\item    We propose a novel on-device pipeline for adding emoji(s) in transcribed text. To the best of our knowledge, this is the first on-device system for emoji addition in transcribed text.
		\item    We train a compact on-device emoji boundary detection model and present experimental evaluations and comparisons.
		\item   We present a compact on-device emoji prediction model and demonstrate competitive results with 80\% fewer parameters.
	\end{itemize}

	\section{Background}
	\subsection{Sentence Boundary Detection}
	Recovering sentence boundaries from speech and its transcripts is a core task for speech recognition as it is essential for readability and downstream tasks. Most research focuses on using either prosodic \cite{haase2001deriving} features or lexical cues \cite{stolcke1998automatic} \cite{batista2012bilingual}, but most state of the art results were achieved using both prosodic and lexical information \cite{gotoh2000sentence}\cite{xu2014deep}. Recent work of \cite{makhija2019transfer} gives the state-of-the-art performance on predicting punctuation. They use transfer learning on pre-trained BERT embeddings \cite{bert} and train a Bi-LSTM with a linear CRF classifier on top of these embeddings. All these researches focus mainly on detecting punctuation in the transcribed text, unlike our task. For adding emoji(s) in transcribed text for emoji prediction, we require detecting sentence sub-parts as well along with sentence boundaries. Moreover, most of these works are server-based and do not focus on latency and memory, two important metrics for on-device performance evaluation. 
	
	\subsection{Emoji Prediction}
	Emoji prediction is an essential feature of almost all modern keyboards. The first modern work of emoji prediction was presented with Deepmoji \cite{felbo2017using} architecture. However, despite its phenomenal accuracy in predicting emoji(s), it is too large to be deployed on-device. More recently, \cite{ramaswamy2019federated} presented a federated learning framework of predicting emoji(s) on-device, with the model trained on user data where they use CIFG. While this helps in user personalization and prediction with out-of-vocabulary (OOV) words, it presents its challenges of network availability, security breaches as showcased by \cite{yang2018federated}.  The state-of-the-art (SOTA) benchmarks  \cite{ma2020emoji} show that BERT-based \cite{devlin2018bert} models largely outperformed the other existing \cite{felbo2017using}  architectures. But these models are not suitable for devices where memory footprint and response time are constrained. Recently there has been ongoing research to reduce the memory footprint of BERT \cite{sun2020mobilebert}, but still, the model size is around 100 MB and hence we do not consider the same for our evaluation. In this paper, we present a memory-efficient emoji prediction model that requires zero network usage and is completely on-device, preventing any data privacy issues. Also, we handle OOV words by the combination of character and word embeddings.

	\section{Processing Pipeline}\label{pipeline}
	Figure~\ref{fig:process pipeline} shows an overview of the processing pipeline. The individual components in the pipeline are discussed in detail below. Since there is no open dataset to evaluate the complete pipeline: Emoji Boundary Detection + Emoji prediction on transcribed text from the audio input, the procedures described in this paper are applied to a set of over 10000 samples collected from 10 individuals' chat annotated by 10 experts from different backgrounds and age groups to include diversity.  These samples being actual user data provide a meaningful evaluation of the solution developed. Apart from this, for training and benchmarking individual modules, various open-source datasets in conjugation with our custom dataset are used, details of which are provided in the respective modules. We considered using the SemEval dataset \cite{barbieri2018semeval}. But this includes tweets that contain one and only one emoji, whereas our main focus is to predict multiple emojis in-between as well.

	\subsection{Text Processing : Emoji Boundary Detection}
	\label{Boundary}
	The emoji boundary detection module identifies where to predict emoji(s) in a given transcribed text. For this purpose, we identify two major categories of boundaries. Emoji can either be predicted at the end of a sentence or within a sentence when a sub-part is detected.
	
\begin{figure}[ht]
		\centering
		\includegraphics[width=\linewidth]{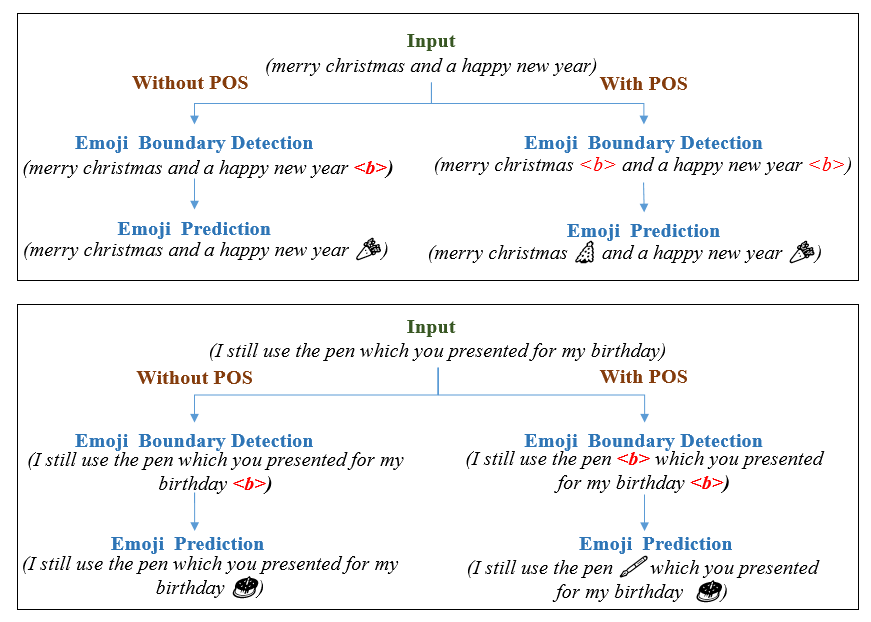}
		\caption{Sample results with and without pre-processing with POS tags}
		\label{fig:POS}
	\end{figure}

	For this task, we use CNN/DM dataset \cite{hermann2015teaching} for training the model. We have chosen this dataset as it has an MIT license, making it ideal for unhindered research. Though we can detect sentence boundaries by identifying punctuations of type `.', `?', etc., but we need to identify sub-parts in a sentence as well since emoji may be needed to be inserted even if there is no sentence boundary (as shown in the examples in Figure~\ref{fig:POS}). We use the Open NLP part-of-speech (POS) tagger \cite{baldridge2005opennlp} to identify these sub-parts.

	Our emoji boundary detection task doesn't require all the labels in a standard POS tagger. Since our task focuses on splitting a sentence into meaningful sub-parts, we experimented with fewer tags without losing the context of the sentence. There is a list of 36 POS tags in the Penn Treebank Project \cite{marcus1993building}. For splitting a sentence into meaningful sub-parts, we recognize and use a subset of these tags for our task. We use CC (coordinating conjunction), IN (subordinating conjunction), WP (Wh-pronoun), WP\$ (Possessive wh-pronoun), and WDT (Wh-determiner) tags.
	
\begin{figure}[ht]
		\centering
		\includegraphics[width=0.7\linewidth]{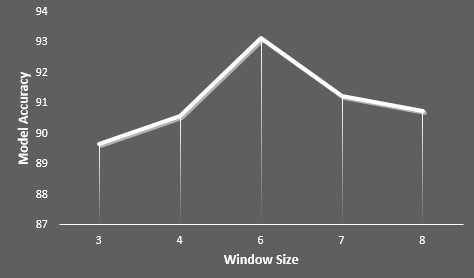}
		\caption{Graph between window size and model accuracy}
		\label{fig:graph}
	\end{figure}

	The POS tagger is then used to prepare the CNN/DM dataset for emoji boundary detection. This involves processing the dataset using a POS tagger to determine sub-parts in the sentence. This is done as we observe that more valid boundaries get detected after using POS tagger in pre-processing as clearly shown in Figure~\ref{fig:POS} and thus improving the user experience significantly. Further, we convert each article from the resulting POS processed dataset to samples of size 6 words, as we found this to give optimal results after thorough experimentation the results of which can be seen in Figure~\ref{fig:graph}. This is done by sliding the window of size 6 by one word at a time. We add padding at the start and the end of the article to make sure the boundary is predicted for the starting and ending words as well. The labels are then determined based on if there is a boundary after the 4th word of each sample. `True' represents the presence of boundary and `False' represents absence.

\begin{figure}[h]
		\centering
		\includegraphics[width=\linewidth]{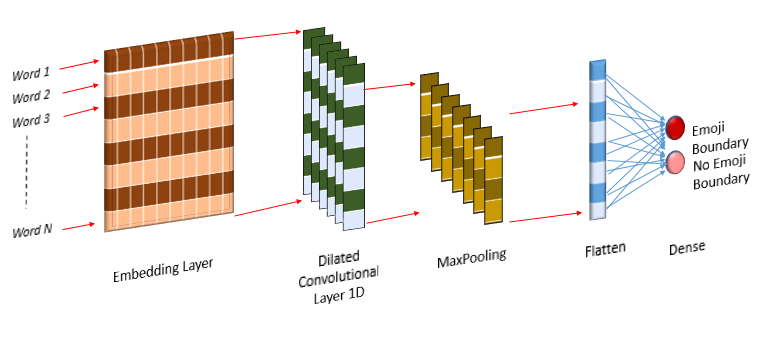}
		\caption{Architecture of Emoji boundary detection.}
		\label{fig:sbd}
	\end{figure}

	We train a CNN-based model on the processed CNN/DM dataset for emoji boundary detection with a vocabulary size of 20K. This model consists of an embedding layer (pre-trained glove embedding), 1D convolutional layer (with 512 filters, kernel size of 3, and dilated by a factor of 2), Max pooling layer with pool size 2, `Flatten' layer which is then followed by a `Dense' layer. We use dilated CNN as they cover a larger receptive field without loss of context, which is faster and also better performing
	\cite{yu2015multi}.  The model architecture is shown in Figure~\ref{fig:sbd}.

	\begin{table*}[ht]
		\caption{Emoji Boundary Model Performance Evaluation}
		\label{t5}
		\centering
		\resizebox{\textwidth}{!}{\begin{tabular}{ c c c c c c c c c }
				\toprule
				\multicolumn{1}{c}{\textbf{Model}} &
				\multicolumn{1}{c}{\textbf{Embedding Dimension}}&
				\multicolumn{1}{c}{\textbf{Class Weight Ratio}}&
				\multicolumn{1}{c}{\textbf{Size}} &
				\multicolumn{1}{c}{\textbf{Precision}} &
				\multicolumn{1}{c}{\textbf{Recall}} &
				\multicolumn{1}{c}{\textbf{F1 score}} &
				\multicolumn{1}{c}{\textbf{Accuracy}} & 
				\multicolumn{1}{c}{\textbf{Multi-line Accuracy}} \\
				\midrule
				CNN &100  & NA  & 2.15MB & 90.77\% &  87.5\% & 89.10\%  & 95.29\%  &  93.37\%  \\
				\rowcolor{Gray}
				CNN & 50 & NA  & 819KB & 90.02\% &  85.75\% & 87.83\%  & 94.7\%  &  92.17\%  \\	
				BiLSTM & 64 & 0.15  & 468KB &  68.3\% & 91.88\%  & 78.36\% & 88.51\% &  84.52\% 	\\		
				BiLSTM & 64 & 0.2  & 468KB &  73.27\% & 89.01\%  & 80.38\% & 90.1\% &  87.86\% 	\\			BiLSTM & 64 & 0.25  & 468KB &  74.21\% & 85.62\%  &  79.509\% & 90.01\% &  86.9\% 	\\			BiLSTM & 64 & 0.35  & 468KB &  62.91\% & 50.04\%  & 55.74\% & 82.01 &  84.48\% 	\\	
				\bottomrule
		\end{tabular}}
	\end{table*}

	\begin{table*}[!ht]
		\caption{Multi-sentence Validation Set Samples}
		\label{t7}
		\centering
		
		\begin{tabular} {p{0.9\linewidth} }
			\toprule
			\multicolumn{1}{c}{\textbf{Input Samples}} \\ 
			\midrule
			I saw one amazing dress today I wanted to buy it so badly but it was so expensive now i regret not getting it \\
			I dont think i can make it I am really sorry I have an important meeting with my client I will definitely come next time \\
			Hey what did you have for lunch today I had ordered pizza it was delicious \\
			Today was a very busy day for me my boss gave me extra work and I had to complete by evening now i feel like going out for a drink would you like to join me \\
			
			\bottomrule
		\end{tabular}
		
	\end{table*}
	
	The model takes input samples consisting of six words. The samples are processed before feeding to the model. The pre-processing involves: converting the words to lowercases, separating short forms like (it’s, it’ll, doesn’t, won’t, we’re, etc.), and removing special characters. This pre-processed data is then sent to the trained model to determine the presence and absence of boundary after each word in the sample. This is based on the output probabilities of the model which is calculated as follows:
	
	\begin{center}
		\begin{equation}
		\begin{multlined}[\columnwidth]\label{eq1} 
		P(boundary~|~i)= sigmoid (y^{out} (i-3, i+2))\\
		i \in (i, words)
		\end{multlined}
		\end{equation}
	\end{center}
	Where $y^{out}$ represents the model output logits for token $i$. The boundary is predicted for the 4th word by considering the previous 3 and the next 2 words to get contextual and semantic information. The output is the probabilities for two labels: boundary present, and not present.

	We use scaled loss function to train our model as follows: 
	\begin{center}
		\begin{equation}
		\begin{multlined}[\columnwidth]\label{eq2}
		Scaled ~Loss = -\sum_{batch size} B  \{log ( y_{B}^{out})\}  + \biggl\{ \frac{N}{M}  (1 - B) \\ (1-y_{B}^{out}) \biggl\}
		\end{multlined}
		\end{equation}
	\end{center}
	Where B is the true label for a boundary, N is the number of samples with boundary, M is the number of samples without boundary, and $y_{B}^{out}$ is the model output probability for a boundary. The training accuracy of the above model is 93.15\% on CNN/DM dataset.


	Table~\ref{t5}  shows the results of evaluating various architectures using the validation set (mentioned in Section \ref{pipeline}) for the task of Emoji Boundary detection. Here, multi-sentence accuracy is the accuracy calculated over test samples consisting of more than one sentence some of which are shown in Table \ref{t7}. Analyzing the results, we use the CNN model with embedding dimension 50 considering model size and performance.
	As compared to the state-of-the-art BERT-based model \cite{makhija2019transfer} which has a model size in hundreds of MB, our model size is just under 1MB.

	\begin{table}[H]
		\caption{Multi-sentence evaluation results on DailyDialog}
		\label{t6}
		\centering
		\begin{tabular}{ c c }
			\toprule
			\multicolumn{1}{c}{\textbf{Evaluation Metric}} & \multicolumn{1}{c}{\textbf{Value (in \%)}} \\ 
			\midrule
			Precision	& 94.307\\
			Recall & 78.673\\
			F1 Score & 86.754 \\
			Accuracy & 94.81 \\
			\bottomrule
		\end{tabular}
	\end{table}
	\begin{table*}[ht]
		\caption{Emoji Prediction Model Performance Evaluation}
		\label{t2}
		\centering
		\begin{tabular}{ c c c c c }
			\toprule		
			\multicolumn{1}{c}{\textbf{Model}} & 
			\multicolumn{1}{c}{\textbf{Top 1 Accuracy}} &
			\multicolumn{1}{c}{\textbf{Top 5 Accuracy}} & 
			\multicolumn{1}{c}{\textbf{F1 score}} & 
			\multicolumn{1}{c}{\textbf{No. of parameters}}\\
			\midrule
			Bag of Words &	12.5\%	& 27\%	& 11.20 &	3.2M\\
			LSTM (Char Embedding) &	17.2\% &	36\% &	14.71 &	0.2M\\
			LSTM (Word Embedding) &	18.7\% &	40.5\% &	15.89 &	4.2M\\
			\rowcolor{Gray}
			ACA LSTM	& 20.1\%	& 42.7\%	& 16.43	& 4.3M\\
			DeepMoji &	22.8\%	& 44\%	& 17.14	& 22.4M\\
			\bottomrule
		\end{tabular}
		
	\end{table*}
	\begin{table*}[ht]
		\caption{On-device model Performance}
		\label{t4}
		\centering
		\resizebox{\textwidth}{!}{\begin{tabular}{ c c c c c }
				\toprule
				\multicolumn{1}{c}{\textbf{Model}} & 
				\multicolumn{1}{c}{\textbf{RAM (MB)}} & 
				\multicolumn{1}{c}{\textbf{ROM (MB)}} & 
				\multicolumn{1}{c}{\textbf{Model Initialization time}} & 
				\multicolumn{1}{c}{\textbf{Inference Time}} \\
				\midrule
				Emoji boundary (CNN)	& 10 & 0.819 & 10ms & 0.4ms/word \\
				Emoji prediction & 3 & 3 & 250ms & 0.8ms/word \\
				Emoji in Dictation Solution (Emoji boundaary (CNN) + Emoji prediction) & 15 & $\sim$4 & 260ms & 1.25ms/word\\
				\bottomrule
		\end{tabular}}
		
	\end{table*}
	
	\begin{table*}[ht]
		\caption{Future Enhancements for VoiceMoji}
		\label{tenh}
		\centering
		\begin{tabular}{p{0.3\linewidth}  p{0.31\linewidth}  p{0.31\linewidth} }
			\toprule
			\multicolumn{1}{c}{\textbf{Input}} & 
			\multicolumn{1}{c}{\textbf{Expected Output}} & 
			\multicolumn{1}{c}{\textbf{Voicemoji Current Output}}  \\
			\midrule
			because it was raining i came late to office
			& because it was raining [rain emoji] i came late to office [office emoji] & because it was raining i came late to office [office emoji] \\
			i came late to office because its was raining
			& 	i came late to office [office emoji] because its was raining [rain emoji] & 	i came late to office [office emoji] because its was raining [rain emoji]  \\
			\midrule 
			do you really think i am happy
			
			& do you really think i am happy [sarcastic emoji] & do you really think i am happy [happy emoji] \\
			\midrule 
			yesterday is mine birthday & yesterday was my birthday [cake emoji] &  yesterday is mine birthday [cake emoji] \\ 
			\bottomrule
		\end{tabular}
		
	\end{table*}

	To further ascertain, the efficacy of our model, we benchmarked on the DailyDialog dataset \cite{li2017dailydialog}, shown in Table~\ref{t6}. We extracted multi-sentence dialogues (dialogues having more than one subpart which is determined by both punctuations and POS tags) specifically, as sentences with just one subpart carry no significance for our use case. Despite the small number of parameters, our model performs well and it is noted that while our model may miss some possible boundaries, it has high precision in the ones that it predicts which is more valuable over recall for our task. Since it's a bad user experience if the user has to correct the system offered predictions.

	To deploy on-device, we convert the frozen model (of size 3.5MB) to tflite using TocoConverter [3] and reduce the size of the model using the OPTIMIZE\_FOR\_SIZE option. This results in a model size reduction of 75\% (Final model size: 819KB). We use this model for on-device inferencing. Standalone execution of this model on-device uses 10MB of RAM and takes an inference time of $\sim$4ms for 10 words.

	\subsection{Text Understanding : Emoji Prediction}

	\label{Prediction}
	After identifying the emoji boundary, we predict emoji(s) for each subpart. We present IdeoNet modeling which deals with semantic analysis of user text to suggest relevant emoji(s). For building semantic IdeoNet, we experimented with different classification models trained on the Twitter corpus \cite{pak2010twitter}.
	
	We explore various deep learning approaches which involve a baseline LSTM model using an attention mechanism with skip connections. Further, we optimize the baseline by analyzing the impact of parameter reduction on model performance and accuracy. However, none of the approaches yielded satisfactory accuracy due to the presence of Out of Vocabulary (OOV) words.

	We propose a novel architecture \textbf{ Attention-based Char Aware (ACA) LSTM} for generating a lightweight model to handle OOV words. This model uses CNN for learning char level embedding such that similar representation is learned for phonetically similar words. Also, both semantic and orthographic information is preserved by aggregating Char-level and Word representation using the attention mechanism.

\begin{figure}[ht]
		\centering
		\includegraphics[width=\linewidth]{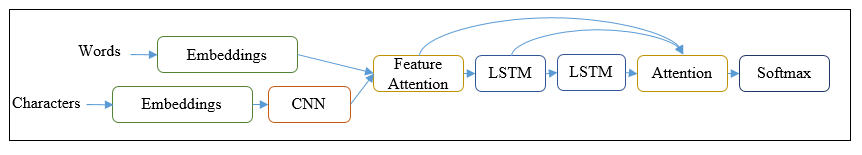}
		\caption{ACA LSTM Architecture}
		\label{fig:aca}
	\end{figure}

	The Model consists of embedding layer - size 16 (custom embedding trained along with the model with the same corpus), Convolutional layer - size 30
	(Kernels: 5,10,15 of width 1,2,3 respectively), Attention Layer - size 46 (word Embedding + CNN), LSTM Layer1 (128 nodes), LSTM Layer2(128 nodes), Attention Layer - size 302 (concatenating output of Attention layer 1, LSTM Layer 1 and 2 ), Softmax class with 64 labels (Frequently used emojis) as illustrated in Figure~\ref{fig:aca}.
	
	We evaluate the approaches discussed above on the Twitter dataset ( $\sim$20K sentences) for the Emoji prediction task. Table~\ref{t2} shows that with ACA LSTM, we achieved comparable accuracy to DeepMoji [2] with an $\sim$80\% lesser parameters.

	\section{Results}
	We evaluate performance both individual component-wise (explained in Section \ref{Boundary} and Section \ref{Prediction}) and of the pipeline.
	
	To validate the complete pipeline, since there is no open dataset for this as explained in Section \ref{pipeline}, we crowdsource the validation data of 50000 chat messages. From this, the messages where emoji is used are retained, reducing the total number of messages to $\sim$10000. Dictation output only presents dictionary words, thus we pre-process the samples collected to remove redundant data and special characters such as digits, \$, etc. Also, we remove code-switched chat messages (that consist of words from non-English language but written using English letters). We expand various chat short forms using a script. For example, “gn” is expanded to “good night”.

	We use the Samsung A50 smartphone model (Android 9.0, 6GB RAM, 64GB ROM, Samsung Exynos 7 Octa9610) to experiment and measure on-device performance, the results of which are shown in Table~\ref{t4}. Overall solution accuracy is calculated based on if our model predicts emoji at the correct boundary and if emoji is predicted from the same category as the one used by the user. The overall accuracy of our solution is 94\%. Voicemoji has an average inference time of 7.2 ms for a 6-word sentence and a total ROM of $\sim$4MB. Voicemoji as a library consumes only $\sim$15 MB RAM.

	Table~\ref{tenh} shows details of future enhancements and some negative scenarios where our proposed system of Voicemoji fails. As clearly shown, the first example shows the system behavior comparison between two similar but rephrased inputs. As observed, in the case of the presence of `because' in the middle, since our sentence boundary prediction model is trained on such scenarios with the help of POS tagging, it gives the correct output for boundaries, but in the first sentence due to sentence starting with `because', it fails to predict a boundary after `raining'. But it is to be noted that this does not result in a completely wrong output and is acceptable. The second scenario is that incase of sarcasm. Since we haven't integrated a sarcasm detection module, our system fails to give correct emoji predictions. This is to be improved as future work for our proposed solution. The last scenario is in the case of grammatical errors. This as well is due to the absence of a grammar correction module, which is to be explored as part of future work along with sarcasm detection. 
	
	
	\section{Conclusion}
	
	We introduce a first-of-its-kind pipeline to insert emoji(s) in dictation in real-time on-device. The initial results of these procedures are promising and demonstrate how dictation can be enriched with emojis and made more fun, thus paving way for a true hands-off dictation experience. This pipeline can be easily extended for other languages, once respective building blocks are adapted for a new language. While evaluating the solution performance, a major category of failure in incorrect emoji prediction is observed due to a lack of sarcasm detection where there is a scope to improve emoji prediction.
	
	\bibliographystyle{IEEEtran}
	
	\bibliography{IEEEexample}
	
\end{document}